\documentclass[12pt]{iopart}
\usepackage{graphicx}
\usepackage{xcolor}
\usepackage{colortbl}

\usepackage{comment}
\begin{document}

\title{Text Classification in Memristor-based Spiking Neural Networks}

\author{Jinqi Huang$^{*1}$, Alex Serb$^2$, Spyros Stathopoulos$^1$, Themis Prodromakis$^2$}
\address{$^1$ Electronics and Computer Science, University of Southampton}
\address{$^2$ School of Engineering, University of Edinburgh}
\ead{$^*$J.Huang@soton.ac.uk}

\vspace{10pt}

\begin{abstract}
Memristors, emerging non-volatile memory devices, have shown promising potential in neuromorphic hardware designs, especially in spiking neural network (SNN) hardware implementation. Memristor-based SNNs have been successfully applied in a wide range of applications, including image classification and pattern recognition. However, implementing memristor-based SNNs in text classification is still under exploration. One of the main reasons is that training memristor-based SNNs for text classification is costly due to the lack of efficient learning rules and memristor non-idealities. To address these issues and accelerate the research of exploring memristor-based spiking neural networks in text classification applications, we develop a simulation framework with a virtual memristor array using an empirical memristor model. We use this framework to demonstrate a sentiment analysis task in the IMDB movie reviews dataset. We take two approaches to obtain trained spiking neural networks with memristor models: 1) by converting a pre-trained artificial neural network (ANN) to a memristor-based SNN, or 2) by training a memristor-based SNN directly. These two approaches can be applied in two scenarios: offline classification and online training. We achieve the classification accuracy of 85.88\% by converting a pre-trained ANN to a memristor-based SNN and 84.86\% by training the memristor-based SNN directly, given that the baseline training accuracy of the equivalent ANN is 86.02\%. We conclude that it is possible to achieve similar classification accuracy in simulation from ANNs to SNNs and from non-memristive synapses to data-driven memristive synapses. We also investigate how global parameters such as spike train length, the read noise, and the weight updating stop conditions affect the neural networks in both approaches. This investigation further indicates that the simulation using statistic memristor models in the two approaches presented by this paper can assist the exploration of memristor-based SNNs in natural language processing tasks.

\end{abstract}

\section{Introduction}

SNNs \cite{SNN}, as the third generation of ANNs \cite{3generation}, have become a widely-researched topic recently. Their spike-based nature and temporal information coding scheme bring advantages in biological plausibility \cite{plausible, plausible2}, sparsity \cite{sparse}, and stochasticity \cite{stochastic}, further making them power-efficient tools compared with traditional ANNs. Therefore, they have been widely used in applications spanning across different areas, such as hand gesture detection \cite{objectDetection, yolo}, gait detection \cite{gait}, financial time series prediction \cite{financial}, music composer classification \cite{composer}, real-time signal processing \cite{signalProcessing}, disease detection \cite{disease-detection}, and robotics \cite{planning,robot,robot_review}. 

Despite the promising potential, hardware implementations of SNNs in conventional Von-Neumann processors can hardly compare with the biological systems regarding energy efficiency. This performance gap mainly stems from the frequent data movements between processing units and the memory to fetch SNN-specific internal variables such as membrane voltages \cite{hardware_requirement,von_neumann_bottleneck}. In-memory computing allows an alternative approach to implementing SNNs in hardware. The computation is executed locally without frequent data movements by combining the processing units and the memory. This working scheme removes the restriction set by the limited communication bandwidth that commonly exists in Von-Neumann processors. An SRAM-based in-memory computing system \cite{sram_in_memory} has reported having the energy efficiency 10x better than google's TPU \cite{tpu}. Based on this idea, many dedicated SNN hardware designs have been developed, from fully digital \cite{loihi, truenorth} to mixed-signal \cite{neurogrid}, from large-scaled computation platforms \cite{spinnaker} to systems-on-wafer \cite{brainscales_soc}. Some designs have been reported to achieve biological speed \cite{truenorth} or multiple orders of magnitude superior to their conventional Von-Neumann predecessors \cite{loihi}.

Meanwhile, the development of emerging non-volatile technology has also seen growth in the past few years. Memristors are known as one of the most mature emerging resistive memory devices among all. The extreme downscaled size \cite{extremeScale}, the large-scale integrability \cite{largeArray}, multi-bit storage \cite{multibit}, crossbar arrangement \cite{1t1r}, ns-scale switching speed \cite{highspeedrram} and high power efficiency \cite{ultralowcurrentrram} make memristors good candidates to replace SRAM in in-memory computing applications \cite{hardwarebayesian, Yao2020, Ambrogio2018}. Moreover, memristors behave similarly to biological synapses in neural networks resulting from their inherent physical properties inspiring their use in spiking neural networks. As examples, References \cite{unsupervised1, unsupervised2} incorporated memristors as synapses in spiking neural networks to perform unsupervised learning using spike-timing-dependent plasticity (STDP) \cite{stdp}. References \cite{online, Boybat2018, Liu2015, Liu2015a} demonstrated pattern recognition and image classification in memristor-based spiking neural networks.

However, only a few works reported memristor-based spiking neural networks for text processing tasks: Design \cite{SNU} employed phase-change memory (PCM), a type of memristor device, to accelerate spiking neural network for language modelling, and Design \cite{onlineTraining} performed a pattern generation task in recurrent spiking neural networks with PCM models using e-prop learning rule \cite{eprop}. One of the main challenges is the costly training of memristor-based spiking neural networks for text processing tasks. Firstly, gradient-based learning rules, regardless of local (e.g. e-prop) or non-local (e.g. surrogate gradient descent \cite{surrogate, surrogateDerivative}), require averaging the accumulated errors over the spike train window that is used to represent a single continuous value. This method considers the effects of every single spike when updating weights. It is inefficient regarding the computational speed and the space efficiency, especially when the length of spike trains for representing a single numerical value is very long and when memristors are involved in the design. Besides, compared to one-hot vectors, word embeddings, such as GloVe \cite{glove} and Word2vec \cite{word2vec}, are more commonly used to represent words in text classification tasks, to improve space efficiency and to map the relations between words. In conventional ANNs, a word embedding layer serves as a lookup table to obtain the dense representation of the input texts. Pre-trained word embeddings are usually fine-tuned later with other network parameters in training to improve classification accuracy. During the training process, they are treated as linear layers using backpropagation (BP) \cite{bp}. However, in SNNs, the dense word representation needs to be converted to spike trains as the input to the network  \cite{neural_word_embeddings, text_spike_rep}. So far, there is no theoretical support regarding how to train word embeddings in spiking neural networks. Lastly, reading and writing memristor arrays can also add variations due to the read noise and the write variation. One solution to these issues is to run the training in a software simulator with realistic memristor models before starting the hardware design. Compared with training in hardware, this solution offers a fast data process and intuition for the system's performance. Reference \cite{neuropack} has introduced an algorithm-level simulator for memristor-based spiking neural networks in memristor models, and Design \cite{online} also used this solution to simulate recurrent spiking neural networks in PCM models. 

In this paper, we develop a simulation framework in Pytorch \cite{pytorch} with an empirical memristor model proposed by \cite{rrammodel} to perform text classification tasks. To use this framework, we firstly present two approaches to obtaining trained memristor-based spiking neural networks for performing text classification tasks. One is converting a pre-trained ANN to its equivalent SNN, and another is training an SNN directly using an ANN-based learning rule. Both approaches train the network using gradient descent-based learning rules without considering the effect of every single spike to improve the computational speed and the space efficiency. In this way, we can also treat a word embedding layer as a linear layer in a conventional ANN and train it using BP. Then, we showcase a sentiment analysis task in the IMDB movie reviews dataset \cite{imdb} to validate both approaches. To the best of our knowledge, this is the first demonstration of a text classification task performed in a spiking neural network with a realistic memristor model. Lastly, we summarise the classification accuracy from both approaches and investigate how global parameters affect the system performance in both approaches.

The contributions of this work are summarised below:
\begin{enumerate}
\item Developing a simulation framework with an empirical memristor model to demonstrate the first text classification task in spiking neural networks with a realistic memristor model.
\item Presenting and benchmarking two approaches to obtain trained memristor-based spiking neural networks for text classification tasks.
\item Investigating system sensitivity to global parameters.
\end{enumerate}

To achieve these goals, we first introduce the overview of the two approaches in section \ref{overview} and explain the details in section \ref{ap1} and \ref{ap2}. Following are the method to add the memristor model in the simulation framework in section \ref{rrammodel} and the weight updating scheme in section \ref{weightupdate}. Section \ref{results} displays all experimental results for the text classification task in the simulation framework with memristor models, and section \ref{conclusion} summarises the paper.

\begin{figure}[t]
    \centering
    \includegraphics[width = 1.0\linewidth]{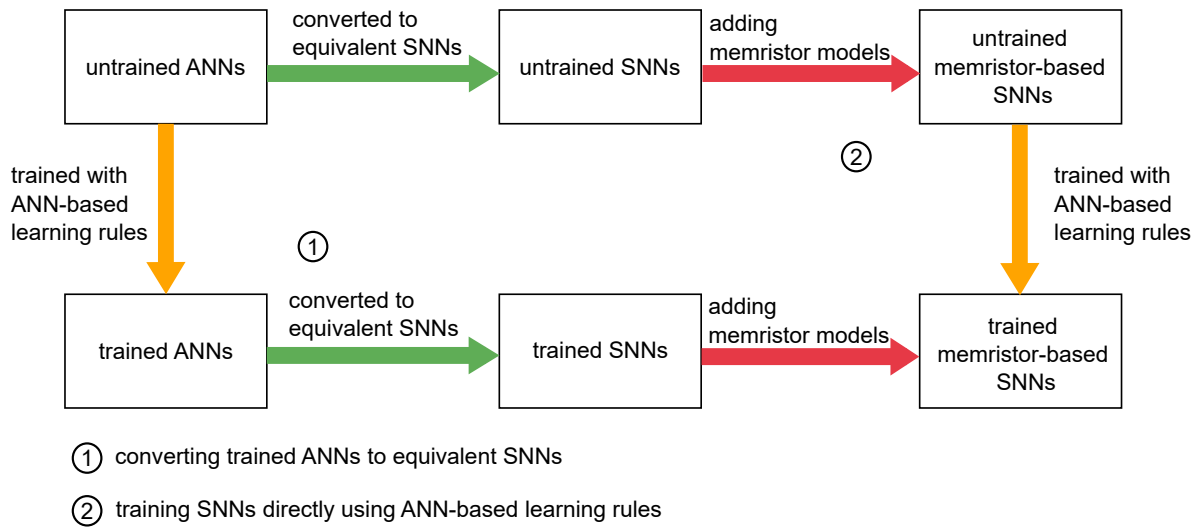}
    \caption{Two approaches to obtaining trained SNNs: approach 1 trains ANNs first (coloured orange), converts trained ANNs to equivalent SNNs (coloured green), and then maps SNN weights to memristors (coloured red); approach 2 converts ANNs to SNNs (coloured green), adds a memristor model (coloured red), and then trains memristor-based SNNs directly using ANN-based learning rules (coloured orange).}
    \label{fig:overview}
\end{figure}

\section{Methods}
\subsection{Methodology Overview}\label{overview}

The main difference between ANNs and SNNs is the conveyance method: ANNs employ continuous values, whereas SNNs use 0-or-1 spikes. As a result, the basic idea to bridge the gap between two neural network architectures is to find the relations to map continuous values to spikes. Reference \cite{anntosnn1} suggested that the spike rates of an SNN are proportional to the ReLU \cite{relu} activation outputs of the ANN equivalents with an error term that is ignorable in shallow networks. Given that inputs and weights are restricted within [0, 1] in a memristor-based SNN, the ReLU activation outputs are always non-negative. Therefore, the spike rates of an SNN are proportional to the input currents. Reference \cite{anntosnn2} further proved that for neurons reset by subtraction, the larger the membrane voltage threshold, the smaller the error term. Additionally, weight normalisation \cite{threhold-balancing} and adaptive threshold \cite{adaptive-threshold} are also commonly used in ANN-to-SNN conversion to match the accuracy. These past works lay the theoretical foundation of the two approaches to obtaining trained memristor-based SNNs proposed in this paper.
 
Figure \ref{fig:overview} gives the diagram of the two approaches: Starting from an untrained ANN, approach 1 trains the ANN using typical ANN-based learning rules, then converts the trained ANN to a trained SNN; approach 2 firstly converts the ANN to its equivalent SNN before training with ANN-based learning rules. After obtaining the trained SNN, memristor models are added to store the weights. In approach 1, this step is as simple as mapping the weights in memristor arrays as conductance; in approach 2, memristor resistance states (RSs) are trained alongside the whole SNN. Therefore, the evolution of RSs can be observed during the training process when using approach 2.

\begin{figure}[t]
    \centering
    \includegraphics[width = 1.0\linewidth]{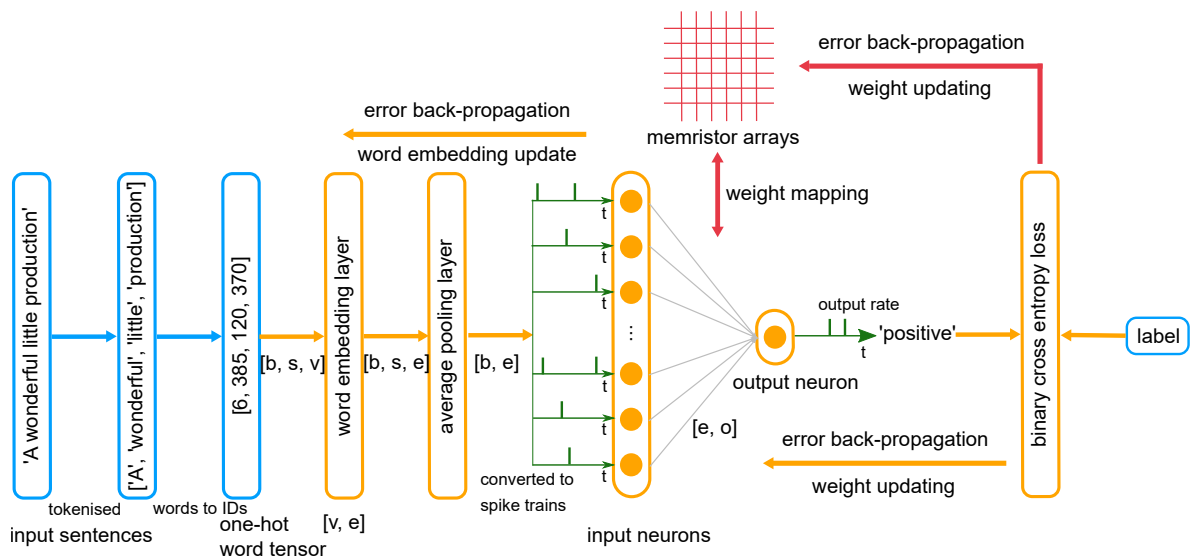}
    \caption{Workflow of the simulation framework for performing the sentiment analysis task in the IMDB review dataset. Inputs, ANN structure, SNN-specified structure, and memristor-related modules are coloured blue, orange, green and red, respectively. $b$, $s$, $e$, and $o$ represent batch size, sentence length, word embedding dimensions, and output dimensions, respectively. We use a movie review, 'A wonderful little production', as an example to explain the input pre-process procedure: firstly, input sentences are tokenised to lists of words ('['A', 'wonderful', 'little', 'production']' in the example). Secondly, the word lists are converted to lists of word IDs ('[6, 385, 120, 370]' in the example). Sentences in the same batch are padded to have the same length. The lists of words are further transformed to a one-hot vector, and padded one-hot vectors of the sentences in the same batch are packed into an input representation tensor with the size of ($b\times s\times v$). Notably, in Pytorch, the word ID lists do not need to be converted to one-hot representation.}
    \label{fig:flow}
\end{figure}

Now we walk through the workflow of the designed simulation framework performing the sentiment analysis task to explain the two approaches. Figure \ref{fig:flow} shows the network we employed in the sentiment analysis task in the IMDB reviews dataset. After being tokenised using the basic English tokeniser and transformed to word IDs, sentences in the same batch are padded to have the same sentence length and packed into an input one-hot representation tensor with the size of ($b\times s\times v$) (where $b$, $s$, $v$ are batch size, sentence length, and vocabulary size, respectively). The one-hot representation tensor is then fed to a word embedding layer to generate a dense word representation whose size is ($b\times s\times e$) (where $e$ is the word embedding dimension), followed by an average pooling layer to obtain an averaged sentence representation across the whole sentence by squeezing the sentence dimension. 

In approach 1, the averaged sentence representation with the size of ($b \times e$) is sent straight to a linear layer whose weights' size is ($e \times o$) (where $o$ is the output dimension). Specifically, there is only one output neuron in the sentiment analysis task because we only have two categories: 'positive' or 'negative'. The binary cross-entropy loss is calculated once the inference results are obtained, and the errors are back-propagated to update weights in the linear and word embedding layers. After the network is trained, the conversion to an SNN starts. Firstly, the averaged sentence representation is transformed to Poisson spike trains which are delivered to the single-layer spiking neural network afterwards. We use the leaky integrate-and-fire neuron model \cite{lif} in this task. Next, the weights in the SNN are mapped in memristor arrays. When executing an inference, weights are loaded from the memristor arrays. The output neuron membrane voltages are calculated according to the mathematical expression of the neuron model, and the firing states are determined by comparing the voltage with the threshold. The spike rates within a time window give the final inference results: if the spike rate is higher than 50\%, it yields a 'positive' result; otherwise, a 'negative' result is given.   

In approach 2, the averaged sentence representation is likewise transformed to Poisson spike trains, and memristor arrays are randomly initialised. After an inference completes, the errors are back-propagated according to the same ANN-based learning rule used in approach 1. The weight changes are converted to the memristive RSs to be used to trigger RS updates. The errors are further back-propagated to the word embedding layer to generate new spike trains in the following iterations.

In the next two subsections, more details that need to be taken into consideration in implementing two approaches in the simulation framework will be introduced.

\subsection{Approach 1: Converting a Pre-trained ANN to an SNN}\label{ap1}
Weights in the linear and word embedding layers need to be constrained in [0, 1] when training the ANN using approach 1. With this restriction, when converting the ANN to the SNN, weights can be represented according to the following equation:
\begin{equation}
    w = \frac{G - G_{min}}{G_{max} - G_{min}}
\end{equation}
Where $G$, $G_{min}$, $G_{max}$ are the current, the minimum, and the maximum memristor conductance, respectively. The averaged sentence representation is also restricted within [0, 1]. Therefore, when converting the ANN to the SNN, spike trains can be generated using the continuous values in the averaged sentence representation as firing rates. Reference \cite{neural_word_embeddings} proposed a method to determine whether to fire a spike at a certain time step to generate a Poisson spike train using the equation shown below:
\begin{equation}\label{eq:RtoW}
    P_t = \cases{1 &if $x_c> x_{random, t}$\\
0&otherwise\\},\quad 0\le t \leq T
\end{equation}
Where $P_t$ gives the spike generation results of the averaged sentence representation in the given time step $t$ in the spike train that uses $T$ time steps to represent a single continuous value, $x_c$ is the continuous value in the averaged sentence representation, and $x_{random, t}$ is the random value generated in the time step $t$ within [0, 1]. A matrix of random values with the size of ($e \times T$) is initialised before the binary word embedding conversion. The averaged sentence representation is then compared to the random-value matrix to generate Poisson spike trains stored in a matrix with the size of ($e \times T$) as inputs of the single-layer SNN.

Because the weights in the linear layer $W_s$ and the averaged sentence representation $x_c$ are both non-negative when training the ANN, the outputs of the linear layer calculated by the equation $y = \sum W_sx_c$ are never negative. A negative constant offset $C$ need to be added to the ANN outputs when using the sigmoid function to map the continuous-valued outputs to probability. If the negative offset $C$ is not used, the non-negative sigmoid function input will result in an output probability never smaller than 50\%. Thus, to make the probability mapping functional, the term passed to the sigmoid function becomes $(y + C)$ instead of $y$. When the term $(y + C)$ equals 0, the inference result is right on the boundary of dividing two categories. Therefore, $C = -\sum \bar{W_s}\bar{x_c} = - 0.5 \times 0.5 \times e$ where $\bar{W_s}$ and $\bar{x_c}$ are both the theoretical averaged values (0.5 and 0.5, respectively). However, when converted to the SNN, the criterion for dividing categories is the output spike rates rather than the sigmoid output. Therefore, a sigmoid function is not used in SNN inference, and no offset is needed in the converted SNN.

The membrane voltage threshold is an extra parameter not used in an ANN. The threshold value needs to be carefully chosen to achieve comparable classification accuracy in the converted SNN. Membrane voltages that are either much higher or much lower than the threshold can cause information loss \cite{adaptive-threshold}. A threshold value 'in the middle' can potentially cause the least information loss. In converted SNNs, the membrane voltages range is known since both weights and word embeddings are constrained within [0, 1]. When all inputs and weights are equal to '1' along the whole spike train, this gives the highest theoretical firing rate. As explained in \cite{anntosnn2}, the output firing rates of an SNN are proportional to the inputs. Therefore, when we keep the weights equal to '1' and change the inputs to half '1' and half '0', the firing rate of the output neuron is supposed to be 50\%—setting the membrane voltage of the output neuron in this scenario as the threshold with the value of $0.5 \times e$ is a good starting point. It guarantees maximum firing when the inputs are maximised and that under any initial conditions, the output neuron will losslessly accumulate membrane potential in a manner commensurate to the strength of the inputs without saturation or clipping. The threshold value needs further fine-tuning based on the approximated value to achieve better performance.

\subsection{Approach 2: Training an SNN directly} \label{ap2}
Training SNNs with ANN-based gradient descent learning rules \cite{GD} can be problematic due to the non-differential property of membrane voltages. A common solution is to employ surrogate derivatives \cite{surrogate, surrogateDerivative} by using straight-through estimators \cite{STEsPresent, STEsPub, binaryactivation}. This work proposed an alternative solution, which stems from ANN-based gradient-based algorithms, to avoid directly dealing with non-differentiable functions. In this work, we use the leaky integrate-and-fire (LIF) neuron model \cite{lif} with reset by subtraction mechanism: 
\begin{eqnarray}
    V_t = V_{t-1} + \sum W_sx_t - Z_{t-1}V_{th} \\
    Z_t = h(V_t - V_{th}) 
\end{eqnarray}
Where $V_t$, $x_t$, and $Z_t$ denote membrane voltages, input spikes and output spikes at time step $t$, respectively. $V_{th}$ is the membrane voltage threshold, $W_s$ represents the weights in the single-layer SNN, and $h(x)$ is the Heaviside step function. For this two-category classification task, we use the binary cross-entropy cost function with a sigmoid layer:
\begin{equation}
    E = -(\hat{y}\log y + (1 - \hat{y})(1 - \log y))
\end{equation}
Where $E$ represents the binary cross-entropy loss, $\hat y$ and $y$ are labels and the probabilities given by the network, respectively. 

As the learning rule, we choose Adaptive Gradient Algorithm (Adagrad) \cite{adagrad}, a gradient-based learning rule with an adaptive learning rate incorporating the past gradients. The learning rule is explained by the equations below:
\begin{eqnarray}
    s \leftarrow  s + g^2 \\
    \theta \leftarrow  \theta - \eta \frac{g}{\sqrt{s} + \epsilon}
\end{eqnarray}
Where $g$ is the gradient of the objective function over some variable of interest $\theta$, $s$ is a state variable that accumulates the square of the gradients, and $\epsilon$ is a small value to avoid dividing by 0.

Now we walk through the steps to acquire the final expression of the weight updates. For the same reason explained in subsection \ref{ap1}, the output neuron firing rate is always non-negative. Therefore, the output firing rate needs to add a negative offset before being passed to a sigmoid function. The offset chosen here is -0.5. The output probabilities of two categories used in calculating the loss are given by passing the output firing rate with a negative offset to a sigmoid function:
\begin{eqnarray}
    a = r + C \\
    y = \frac{1}{1 + e^{-a}}
\end{eqnarray}
Where $r$ is the output neuron firing rate, $C$ is the negative offset, and $a$ is a temporary variable to save the sum of the firing rate and the offset. 
The output firing rates of an SNN are proportional to the equivalent ANN activation outputs in the equivalent ANN, as proven in \cite{anntosnn2}. Given that the outputs are always non-negative, we further proved that in a memristor-based SNN, the output firing rates are proportional to the equivalent ANN outputs as follows (please see the derivation in supplementary material section 1):
\begin{equation}\label{eq:anntosnn}
    r = \frac{V_c}{V_{th}} - \frac{V_t - V_0}{V_{th} \cdot T}
\end{equation}
Suppose the equivalent ANN has the mathematical expression shown below:
\begin{equation}
    V_c = \sum W_sx_c 
\end{equation}
Where $V_c$ and $x_c$ are continuous-valued ANN outputs and inputs, and $V_0$ is the initial membrane voltage of a spiking neuron when processing a new sample. $V_0$ is 0 in this work because we reset the membrane voltage accumulator before a new inference. The error term $V_t/(V_{th} \cdot T)$ in Equation \ref{eq:anntosnn} depends on the final membrane voltage $V_t$ at the end of the spike window $T$, and $V_c$ is the continuous value that the spike train in the spike window $T$ represents. Intuitively speaking, $V_t$ implicitly depends on $V_c$. Therefore, the error term $V_t/(V_{th} \cdot T)$ is weakly correlated to $V_c$, and this weak correlation can potentially introduce an error in the conversion of an ANN into an SNN. This error can be accumulated in a deep neural network, but it is ignorable in a shallow network \cite{anntosnn2}. If we ignore the weak dependency of the error term to $V_c$ and use two constants $\alpha$ and $\beta$ to replace the error term and the constant-coefficient $1/V_{th}$, we can describe the output firing rates differently:
\begin{equation}
    r = \alpha V_c + \beta
\end{equation}

Therefore, the derivative of $r_t$ with respect to $V_c$ equals to $dr/dV_c = \alpha$. As a result, the gradient of the cost function over the weights in the single-layer SNN is expressed as follows using the chain rule:
\begin{equation}
    \eqalign{\frac{dE}{dW_s} &= \frac{dE}{da}\cdot \frac{da}{dV_c} \cdot \frac{dV_c}{dW_s} \cr
    &= (y - \hat y) \cdot \alpha \cdot x_c}
\end{equation}

Similarly, the word embedding matrix can be updated by further back-propagating the errors:
\begin{equation}\label{eq:gradient}
    \eqalign{\frac{dE}{dW_e} &= \frac{dE}{da}\cdot \frac{da}{dV_c} \cdot \frac{dV_c}{dx_c} \cdot\frac{dx_c}{dW_e} \cr
    &= (y - \hat y) \cdot \alpha \cdot W_s x_e}
\end{equation}
Where $W_s$ denotes the parameters in the word embedding matrix, and $x_e$ is the one-hot input word representation.

In Adagrad, the change amounts are decided by the term $g/(\sqrt{s} + \epsilon)$. Both $g$ and $\sqrt{s}$ are proportional to $\alpha$. Therefore, $\alpha$ can be cancelled out in the expression. The final expression to be used in Adagrad without $\alpha$ is simplified as follows:
\begin{eqnarray}\label{eq:final}
    g_s = (y - \hat y) \cdot x_c \\
    g_e = (y - \hat y) \cdot W_sx_e \label{eq:final2}
\end{eqnarray}

Compared to accumulating errors through the whole spike train using surrogate gradient descent, which can be expressed as the equation below, our method improves the computational speed and the space efficiency by skipping the step of accumulating errors across the spike window.
\begin{equation}\label{eq:surrogate}
    \eqalign{\frac{dE}{dW_s} &= \frac{dE}{da}\cdot \frac{da}{dZ_t} \cdot \frac{dZ_t}{dy_t}\cdot \frac{dV_t}{dW_s} \cr
    &= (y - \hat y) \cdot \frac{1}{T}\sum^{T} h'(V_t - V_{th}) x_t}
\end{equation}

If using a binary activation \cite{binaryactivation} shown in Equation \ref{eq:binaryactivation} to replace the Heaviside step function and choosing a threshold equal or larger than the half of the maximum possible membrane voltage to avoid information loss, the gradient calculated by surrogate gradient descent (Equation \ref{eq:surrogate}) can be simplified to $(y - \hat y) \cdot x_c / (2V_{th})$. Cancelling out the constant coefficient $1/(2V_{th})$ gives the same simplified gradient expressions shown in Equation \ref{eq:final} and \ref{eq:final2}. 
\begin{equation}\label{eq:binaryactivation}
    h'(V_t - V_{th}) = \cases{\frac{1}{2V_{th}} & if $0 < V_t < 2V_{th}$\\
    0 & otherwise\\
    }
\end{equation}

\subsection{Adding Memristor Models} \label{rrammodel}
This work employs the statistical memristor model proposed by \cite{rrammodel} to predict the behaviour of memristors during the training or inference process in memristor-based SNNs. The switching dynamics of memristor devices are shown below for convenience:
\begin{equation}\label{eq:rrammodel}
    \frac{dR}{dt} = \cases{A_p(-1 + \exp(|v|/{t_p}))h(r_p(v) - R)(r_p(v) - R)^2 & if $v > 0$ \\
    A_n(-1 + \exp(|v|/{t_n}))h(R - r_n(v))(R - r_n(v))^2 & otherwise \\
    } 
\end{equation}
Where $A_{p,n}$ denotes the scaling factor, the term $(-1 + \exp(|v|/{t_{p, n}}))$ reflects the exponential increment between the switching rate and the bias voltage $v$, and the rest of the equation portrays the dependency on the current memristive RSs and the dynamic upper/lower boundaries $r_{p, n}$ represented as the equation below with fitting parameters $a_{0p, n}$ and $a_{1p, n}$:
\begin{equation}
    r_{p, n} = a_{0p, n} + a_{1p, n}v
\end{equation}

In practice, all parameters can be extracted using the method proposed by \cite{extraction} with the aid of characterisation tools \cite{arc1, arc2}. This work takes these parameters to predict the memristive RSs given the current memristive RSs and the bias voltages. Using Pytorch, all memristive RSs are updated and accessed simultaneously. Given the pulse voltage \textbf{v} and the pulse-width \textbf{pw}, the duration is converted to iteration times given by $\mathbf{N} = \mathbf{int}(\mathbf{pw} / dt)$ and memristive RSs are updated in loops with equation $\Delta R = \sum\limits_{}^N (dR/dt) $.

\subsection{Weight Updating Scheme} \label{weightupdate}
After obtaining the expected weights to map in memristor arrays, the essential step is to convert the expected weights to expected memristive RSs. Rewriting Equation \ref{eq:RtoW} gives the expression of the expected memristive RSs given the expected weights:
\begin{equation}
    R = \frac{1}{w(\frac{1}{R_{min}} - \frac{1}{R_{max}}) + \frac{1}{R_{max}}}
\end{equation}
Where $R$ denotes the expected memristive RS, and $R_{min}$ and $R_{max}$ are the lower and the upper RS boundaries that are the minimal/maximal of a set of $r_{p, n}$ values with different bias voltage $v$. As shown in Equation \ref{eq:rrammodel}, the switching dynamics of memristor devices are non-linear and governed both by bias voltages and the current memristive RSs. Therefore, it is challenging to trigger memristors to reach certain RSs. Reference \cite{neuropack} proposed a 'predict-write-verify' loop to find the parameter sets (including bias voltages and the pulse-width) that can lead to memristive RSs closest to expected. Given multiple sets of bias voltages and the pulse-width, firstly, predict all resulting memristor RSs, and find the one that contributes to the shortest distance to the expected RSs. Next, apply the pulse with this chosen parameter set, and verify the actual new RSs. Repeat the same procedure until the differences between the actual RSs and the expected RSs are within an acceptable range defined by parameter $ \mathbf{R\ tolerance} = |R_{expected} - R_{actual}| / R_{expected}$. Considering a possible scenario of infinite loops caused by the read noise, another parameter \textbf{maxN} is also supplied to restrict the maximum iteration times.  

When implementing this weight updating scheme, a critical design choice is to choose pulsing parameter sets. Suppose the provided pulsing parameter can only lead to small changes in RSs. In that case, it will take multiple steps to reach the expected RSs and potentially lead to a significant error when reaching the maximum iteration times set by the \textbf{maxN}. If parameter sets only allow large-step RS changes, the targeted RSs may be 'missed'. Therefore, a practical solution is to match the required RS changes and the estimated RS changes that the selected parameter sets can result in. For approach 1, the required RS changes usually are relatively large. Therefore, parameter sets that can lead to large-step and small-step RS changes are both needed: the former is used to make RSs converge quickly, and the latter is for fine-tuning. Approach 2, however, only needs parameter sets that lead to relatively smaller-step RS changes because the required RS changes in each training step are relatively small. To make sure the weight updating can meet the stop condition set by the \textbf{R tolerance}, for both approaches, parameter sets that can lead to RS update smaller than the RS margin defined by the \textbf{R tolerance} need to be chosen.

\section{Results}\label{results}

\subsection{Experiment Overview}

\begin{figure}[t]
    \centering
    \includegraphics[width =1.0\linewidth]{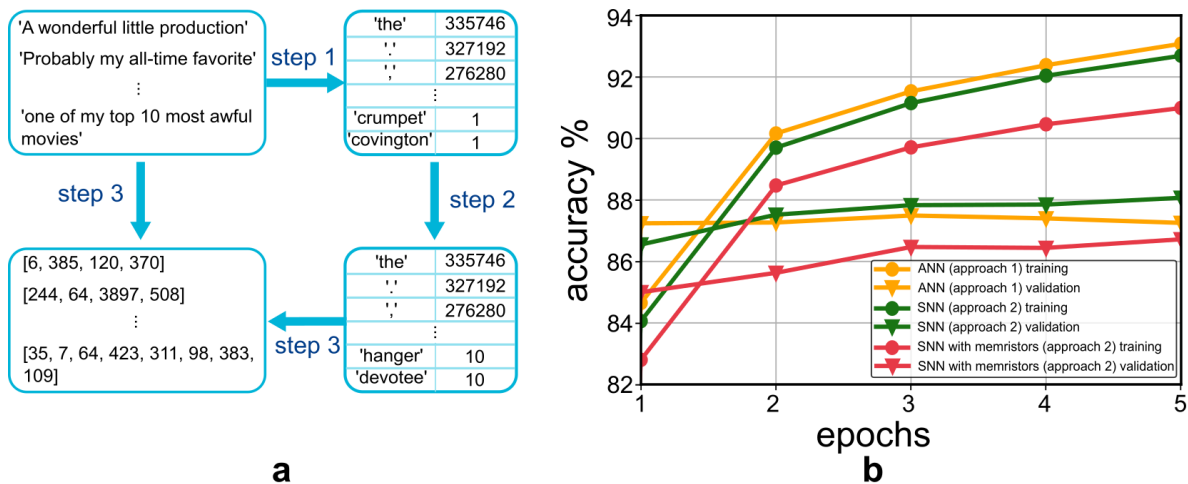}
    \caption{(a) Building a vocabulary using the training samples. Step 1: Create a look-up table for storing word frequency. Step 2: Omit the words that appear less than 10x to reduce the vocabulary size. Step 3: Iterate through all samples to transform them into word IDs using the vocabulary. Words are tagged as '[unk]' if they cannot be found in the vocabulary. When packed into batches, samples in the same batch are padded with the '[pad]' token to have the same sentence length. (b) Training and validation accuracy evolution curves in the ANN (approach 1), the SNN (approach 2), and the SNN with memristor models (approach 2).}
    \label{fig:overall}
\end{figure}

We demonstrate a sentiment analysis task with the IMDB movie reviews dataset to validate the two approaches to obtaining a trained memristor-based SNN. The dataset includes 25k highly polar movie reviews training samples and 25k test samples. Inspired by the GitHub project \cite{sentiment-analysis}, we firstly use the training samples to build a vocabulary to pre-process the input data (Figure \ref{fig:overall} (a)). We count the word frequency in the training samples and put those words that appear over 10x times into the vocabulary. This step creates a vocabulary of 20473 words. Next, we split the original training set into a sub-training set with 17.5k samples and a validation set with 7.5k samples. After that, an input review is tokenised using the basic English tokeniser and further converted into a word ID vector. The one-hot representation of the word ID vector is then fed to the neural network to start an inference. The word embedding layer contains pre-trained GloVe word embeddings \cite{glove} with the embedding dimension of 100. For the training and the inference processes, please refer to Section 'Methods'. In this experiment, the network is trained for 5 epochs. For a single training epoch, after training using the entire training set with 17.5k samples, the training is switched off, and 7.5k samples from the validation set are fed into the network to validate the training effect for this epoch. The training and the validation set samples are shuffled before starting another epoch. The trainable network parameters that lead to the smallest validation loss are reloaded to the network for testing after training. 25k samples from an independent test set are then sent to the network. Experiment parameters for the two approaches are summarised in Table \ref{tab:param}, accuracy evolutions during the training process in Figure \ref{fig:overall} (b), and the final test accuracy in Table \ref{tab:test_accu}. The baseline test accuracy is 86.02\%, given by the ANN in approach 1. According to \cite{sentiment-analysis}, to achieve higher classification accuracy, more complex architectures are needed such as transformers \cite{transformers}. We can notice from Table \ref{tab:test_accu} that, in approach 1, the test accuracy degradations from the ANN to the SNN and the SNN to the SNN with memristor models are very small (0.13\% and 0.01\% respectively). In approach 2, the degradation from the ANN to the SNN is 0.1\%, whilst the test accuracy drops 1.06\% by adding memristor models. 

\begin{table}[h]
    \centering
    \caption{Test accuracy for approach 1 and 2.}
    \begin{indented}
    \item[]\begin{tabular}{@{}cccccc}
    \br
         & ANN & SNN & SNN with memristors& SNN & SNN with memristors\\
    approach & 1 & 1 & 1 & 2 & 2 \\   
    \mr
    accuracy (\%) & 86.02 & 85.89& 85.88& 85.92& 84.86\\ 
    \mr
    \end{tabular}
    \end{indented}
    \label{tab:test_accu}
\end{table}

\begin{table}[h]
    \centering
    \caption{Parameters used in the sentiment analysis tasks baseline configuration for Approach 1 and 2. Different parameter values for different approaches are highlighted in light grey.}
    \begin{indented}
    \item[]\begin{tabular}{@{}ccc}
    \br
    Parameters & Approach 1 & Approach 2\\
    \mr
    Training set size & 17.5k & 17.5k \\
    Validation set size & 7.5k & 7.5k \\
    Test set size & 25k & 25k \\
    Vocabulary size & 20473 & 20473 \\
    Embedding dimension & 100 & 100 \\
    Output dimension & 1 & 1 \\
    Batch size & 1 & 1\\
    \rowcolor[gray]{.8}Offset & -25 & -0.5\\
    T & 1k & 1k \\
    \rowcolor[gray]{.8}$V_{th}$ & 50 & 56.75\\
    $\eta$ & 0.05 & 0.05 \\
    $\epsilon$ & $1\times 10^{-8}$ & $1\times 10^{-8}$\\
    Array size & 10 $\times$ 10 & 10 $\times$ 10\\
    $A_p$ & 0.21389 & 0.21389 \\
    $A_n$ & -0.81302 & -0.81302 \\
    $a_{0p}$ & 37087 & 37087 \\
    $a_{0n}$ & 43430 & 43430 \\
    $a_{1p}$ & -20193 & -20193 \\
    $a_{1n}$ & 34333 & 34333\\
    $t_p$ & 1.6591 & 1.6591 \\
    $t_n$ & 1.5148 & 1.5148 \\
    Positive pulse magnitude (V) & 0.9 & 0.9\\
    \rowcolor[gray]{.8}Positive pulse duration (us)& 1, 2, 10, 20, 50, 100 & 1, 2, 10, 20, 50\\
    Negative pulse magnitude (V) & -1.2 & -1.2\\
    \rowcolor[gray]{.8}Negative pulse duration (us) & 1, 2, 10, 20, 100, 1000, 2000, 5000  & 1, 2, 10, 20, 100\\
    dt (ms) & 1 & 1 \\
    \textbf{R tolerance} & 0.05\% & 0.05\% \\
    MaxN & 5 & 5 \\
    \mr
    \end{tabular}
    \end{indented}
    \label{tab:param}
\end{table}

\subsection{Results Analysis of Approach 1}

\begin{figure}
    \centering
    \includegraphics[width=1.0\linewidth]{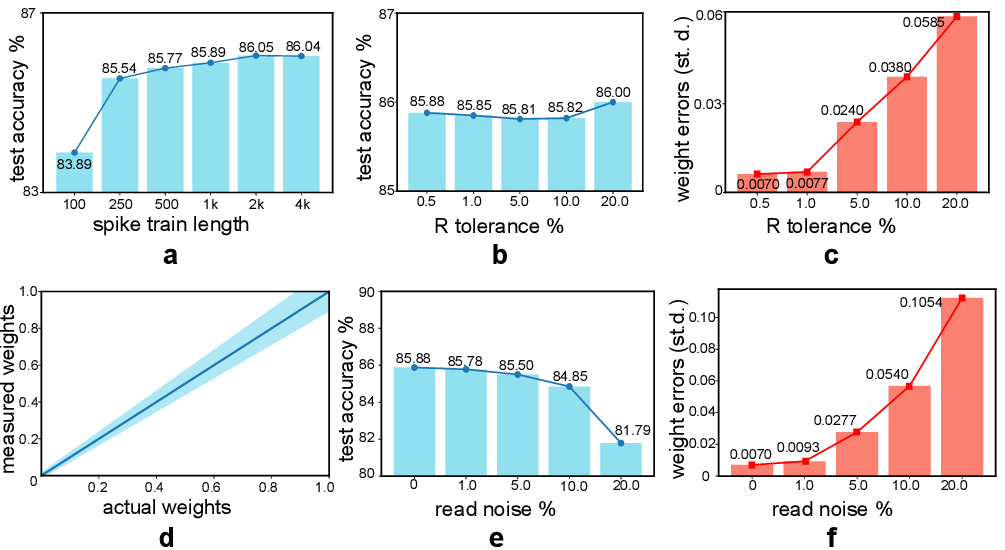}
    \caption{Result analysis for converting an ANN to a memristor-based SNN. (a) Test accuracy vs spike train length, where a train represents a single continuous value. (b) Test accuracy vs the \textbf{R tolerance} values when mapping weights to memristor RSs. (c) Weight standard deviations vs \textbf{R tolerance} when mapping weights to memristor RSs. (d) The measured weights along with the actual weights with the read noise of 10\%. (e) The test accuracy dependency on the read noise. (f) Weight standard deviations vs read noise. }
    \label{fig:ap1}
\end{figure}

Now we look into the results from approach 1. Figure \ref{fig:ap1} (a) shows how the spike train length for representing a single numerical value affects the classification accuracy. From the curve, we can see that the accuracy shows an 'increased-and-saturated' tendency along with the increase of the spike train length because the spike trains are stochastic. The longer the spike trains, the more accurately they can represent the numerical values. However, increasing the length will not change the spike representation accuracy when the spike train is long enough. Therefore, system performance plateaus. Longer spike trains also require longer runtime to iterate through the whole spike train for processing a single sample. Therefore, we choose T = 1k for experiments as the standard configuration in this section, though longer spike trains can result in more accurate representations.

System sensitivity to the \textbf{R tolerance} when using approach 1 is given in Figure \ref{fig:ap1} (b). In theory, the larger the \textbf{R tolerance}, the larger the maximum errors between the expected memristor RSs and the actual values. Therefore, a large \textbf{R tolerance} value could lead to performance degradation. However, as shown in the bar chart, the system is robust to large \textbf{R tolerance} values. To investigate the reasons, we plot the Euclidean distance between the expected and the actual weights along with the increase of the read noise (Figure \ref{fig:ap1} (c)). Even when the \textbf{R tolerance} is as large as 20\%, the Euclidean distance is not higher than 0.06. This indicates that the system can tolerate variations in exporting weights into memristor arrays when converting a trained ANN to an SNN with memristors.

Similarly, approach 1 also exhibits robustness to the read noise of the read operation for memristor arrays (Figure \ref{fig:ap1} (e)). In practice, the instruments used to read memristive RSs introduce read noise, potentially affecting system performance. The read noise in the framework is simulated by adding a random number that does not exceed a certain percentage of the memristor RSs. In other words, when we say the read noise is 10\%, it means that the errors caused by the read noise are no more than $\pm$10\% of the memristor RSs (Figure \ref{fig:ap1} (d)). The read noise is both included in the prediction stage of the weight updating scheme and the inference. In this simulation, even with 20\% read noise, which means the error between the memristor RSs shown by the instruments and the actual values is up to 20\% in the worst case, the system can still achieve a classification accuracy of 81.79\%. Figure \ref{fig:ap1} (f) shows the Euclidean distance between the actual and the measured weights with different read noise values. The Euclidean distance caused by 20\% read noise is around 0.105 (Figure \ref{fig:ap1} (f)). 

\subsection{Results Analysis of Approach 2}

\begin{figure}[h]
    \centering
    \includegraphics[width = 1\linewidth]{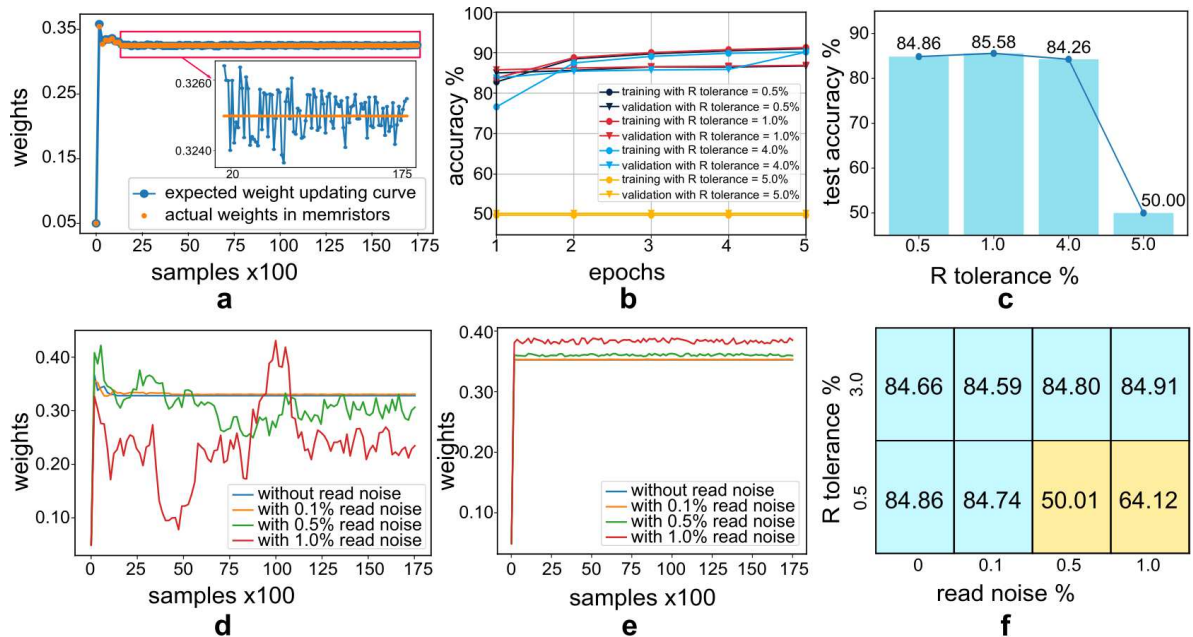}
    \caption{Result analysis for training the memristor-based SNN directly. (a) Weight evolution curve for synapse 0 in the single-layer SNN during training epoch 1. The curve is plotted every 175 samples. The calculated expected weights are plotted in blue, and the actual weights mapped into memristor arrays are in orange. Notably, memristive devices stop updating once the requested updates drop below the \textbf{R tolerance}. (b)-(c) Training/validation and test accuracy with different \textbf{R tolerance} values. (d)-(e) Measured weight evolution for synapse 0 in the training epoch 1 with different read noise values when the \textbf{R tolerance} is 0.5\% and 3.0\%. The curve in blue is the baseline without the read noise. Weights are plotted every 175 samples. (f) Shmoo plot for the test accuracy (\%) with different read noise values (x axis) and different \textbf{R tolerance} (y axis). Positive and negative results are coloured blue and yellow, respectively.}
    \label{fig:ap2}
\end{figure}

Now we investigate how the \textbf{R tolerance} affects weight updating when using approach 2. Figure \ref{fig:ap2} (a) plots the expected weight values evolution in the synapse 0 in epoch 1 when training the memristor-based SNN directly (in the blue curve) and the actual weights written in memristors (as orange dots). The curves are plotted every 175 samples. The expected weight (in the blue curve) progressively converges to the optima. However, the actual weights written in memristors stop updating after a certain point because the weight updating is cut off by the \textbf {R tolerance} when the weight change is smaller than the value set by the \textbf{R tolerance}. We envisage that the larger the \textbf{R tolerance} value, the earlier the weight updating stops. Figure \ref{ap2} (b) and (c) give the training/validation accuracy curves and the final test accuracy with different \textbf{R tolerance} values. The accuracy shows no noticeable change when the \textbf{R tolerance} is smaller than 4.0\%. However, when the \textbf{R tolerance} is increased to 5\%, the network cannot give reasonable classification predictions due to the early cut-off.

Contrary to the read noise-robustness of the system built using the ANN conversion, directly training the memristor-based SNN is sensitive to the read noise. Figure \ref{fig:ap2} (d) depicts the measured weight evolution in synapse 0 during the training epoch 1 with different read noise values when the \textbf{R tolerance} is 0.5\%. Weights are plotted every 175 samples. The ideal situation gives the baseline without the read noise in blue. The weight in the baseline converges to an optimal state with the value around 0.328, and it converges to around 0.330 with 0.1\% read noise. However, the weight evolutions during epoch 1 with the read noise higher than the \textbf{R tolerance} (specifically, with 0.5\% and 1.0\% read noise) fail to converge. We speculate this is because weights cannot be correctly updated when the variations introduced by the read noise exceed the acceptable range defined by the \textbf{R tolerance}. The effects can be multi-folded: the relatively large measured errors confuse the weight updating scheme to determine the weight change magnitudes or even directions, introduce weight wobblings, and make the \textbf{R tolerance} unable to be used as an effective stop condition. With 3.0\% of \textbf{R tolerance}, which is higher than the largest read noise we have tested, weights in all conditions converge (shown in Figure \ref{fig:ap2} (e)). When the \textbf{R tolerance} is higher than the read noise, the major errors during the weight updating are introduced by the \textbf{R tolerance} rather than the read noise so that the impact of the read noise can be ignored. Two features can be noticed in Figure \ref{fig:ap2} (e)): firstly, more significant wobblings exist along with the increase of the read noise, with the largest wobbling of $\sim \pm0.005$; secondly, the larger the read noise, the farther the measured convergent state apart from the baseline, with the largest distance of 0.03. However, both the fluctuations and the distances seem to be within acceptable ranges. Figure \ref{fig:ap2} (f) gives the test accuracy with different read noise and \textbf{R tolerance}. We conclude that when the \textbf{R tolerance} is higher than the read noise, weights converge, and the test accuracy shows no noticeable difference from the baseline; otherwise, weights fail to converge, and the final test accuracy shows negative results.

\section{Conclusion}\label{conclusion}

In this paper, we developed a simulation framework incorporating an empirical memristor model to demonstrate the first text classification task in memristor-based spiking neural networks with the IMDB movie reviews dataset. We took two paths to obtain trained memristor-based spiking neural networks by 1) converting a pre-trained ANN to a memristor-based SNN or 2) training a memristor-based SNN directly using the equivalent ANN-based learning rules. We elaborated on the specific details of estimating the critical parameters to guide users to achieve comparable performance when taking the presented methods. We achieved the test accuracy of 85.88\% and 84.86\% from the two approaches with only 0.14\% and 1.16\% degradations, respectively, given the baseline accuracy of 86.02\% from the equivalent ANN. Finally, we investigated how the SNN stochasticity and memristor non-idealities affect the system performance. We concluded that it is possible to achieve very similar performance in simulation from ANNs to memristor-based SNNs and from non-memristive synapses to data-driven virtual memristive synapses. Proven by demonstrating a sentiment analysis task, using the simulation framework presented in this paper can accelerate research in employing memristor-based SNNs in text classification tasks by overcoming the lack of learning rules and memristor non-idealities.

\section*{Conflict of Interest Statement}
The authors declare that the research was conducted in the absence of any commercial or financial relationships that could be construed as a potential conflict of interest.

\section*{Data Availability Statement}
The original code of this work can be found in https://github.com/hjq310/text-classification-in-memristorsnn.

\section*{References}
\bibliographystyle{unsrt}
\bibliography{bib}

\end{document}


\title{Supplementary Material: Text Classification in Memristor-based Spiking Neural Networks}
\section{Relation between the memristor-based SNN firing rates and the equivalent-ANN outputs}
Inspired by the derivation given by this work \footnote{Rueckauer, Bodo \& Lungu, Iulia-Alexandra \& Hu, Yuhuang \& Pfeiffer, Michael \& Liu, Shih-Chii. (2017). Conversion of Continuous-Valued Deep Networks to Efficient Event-Driven Networks for Image Classification. Frontiers in Neuroscience. 11. 10.3389/fnins.2017.00682. }, now we derive the relation between the memristor-based SNN output firing rates and the equivalent-ANN outputs.

Assume we have an SNN with the integrate-and-fire neuron model described as follows:
\begin{eqnarray}\label{eq:snn}
    V_t = V_{t-1} + \sum W_sx_t - Z_{t-1}V_{th} \\
    Z_t = h(V_t - V_{th}) 
\end{eqnarray}
Where $V_t$, $x_t$, and $Z_t$ denote the membrane voltages, input spikes, and output spikes at time step $t$, $V_{th}$ is the membrane voltage threshold, $W_s$ represents the weights in the single-layer SNN, and $h(x)$ is the Heaviside step function. We also have its equivalent ANN described using the equation below:
\begin{equation}\label{eq:ann}
    V_c = \sum W_sx_c
\end{equation}
Where $V_c$ and $x_c$ are the continuous-valued ANN outputs and inputs. $x_t$ is the spike train representation of $x_c$, and their relation can be expressed using the equation below:
\begin{equation}\label{eq:input}
    x_c = \frac{1}{T}\sum^T x_t
\end{equation}
Where $T$ is the length of the spike train to represent a single continuous value. 

Now we accumulate the membrane voltage $V_t$ over the time window $T$ using Equation \ref{eq:snn}, rearrange it and use Equation \ref{eq:ann} and \ref{eq:input} to simplify it, we can write the expression of $\sum^T Z_{t-1}$:
\begin{equation}
    \eqalign{\sum^T V_t &= \sum^T V_{t-1} + \sum^T\sum W_sx_t - \sum^T Z_{t-1}V_{th}\\
    V_t - V_0 &= \sum W_s \sum^T x_t - V_{th}\sum^T Z_{t-1}\\
    V_t - V_0 &= T\sum W_sx_c - V_{th}\sum^T Z_{t-1}\\
    V_t - V_0 &= T \cdot V_c -  V_{th}\sum^T Z_{t-1}\\
    \sum^T Z_{t-1} & = \frac{T \cdot V_c}{V_{th}} - \frac{V_t - V_0}{V_{th}}}
\end{equation}
By dividing by $T$, we can get the approximation of the firing rate as follows:
\begin{equation}
    r = \frac{\sum\limits_{}^T Z_t}{T} \approx \frac{\sum\limits_{}^T Z_{t-1}}{T} = \frac{V_c}{V_{th} } - \frac{V_t - V_0}{V_{th} \cdot T}
\end{equation}
Where $V_0$ denotes the initial membrane voltage of a spiking neuron when starting a new inference. From the equation, we can see that the memristor-based SNN output firing rates are proportional to the equivalent ANN continuous-valued outputs $V_c$, with a constant coefficient $V_c/V_{th}$ and an error term $(V_t - V_0)/(V_{th} \cdot T)$.